\documentclass[sigconf]{acmart}
\usepackage{caption}
\usepackage{subcaption}
\usepackage{graphicx}
\usepackage{booktabs}
\usepackage{siunitx}
\usepackage{pgfplotstable}
\usepackage{blindtext}
\usepackage[inline]{enumitem}
\usepackage{xcolor}
\usepackage{bm}

\setcopyright{rightsretained}
\acmDOI{10.475/123_4}

\acmISBN{123-4567-24-567/08/06}

\copyrightyear{2017} 
\acmYear{2017} 
\setcopyright{acmcopyright}
\acmConference{GECCO '17}{July 15-19, 2017}{Berlin, Germany}\acmPrice{15.00}\acmDOI{http://dx.doi.org/10.1145/3071178.3071192}
\acmISBN{978-1-4503-4920-8/17/07}

\begin{document}
\title{The Evolution of Neural Network-Based Chart Patterns: A Preliminary Study}
%\titlenote{Produces the permission block, and copyright information}
%\subtitle{Extended Abstract}
%\subtitlenote{The full version of the author's guide is available as
  %\texttt{acmart.pdf} document}

%\author{\  }
%\affiliation{%
    %\institution{\ }
    %\institution{\ }
    %\institution{\ }
    %\institution{\ }
  %\streetaddress{\ }
  %\city{\ }
%}

\author{Myoung Hoon Ha}
%\authornote{Dr.~Trovato insisted his name be first.}
\affiliation{%
    \institution{School of Computer Science \& Engineering}
    \institution{Seoul National University}
  \streetaddress{1 Gwanak-ro, Gwanak-gu}
  \city{Seoul}
  \state{Korea}
  \postcode{155-744}
}
\email{mh.ha.soar@gmail.com}

\author{Byung-Ro Moon}
%\authornote{Dr.~Trovato insisted his name be first.}
\affiliation{%
    \institution{School of Computer Science \& Engineering}
    \institution{Seoul National University}
  \streetaddress{1 Gwanak-ro, Gwanak-gu}
  \city{Seoul}
  \state{Korea}
  \postcode{155-744}
}
\email{moon@snu.ac.kr}

%\renewcommand{\shortauthors}{B. Trovato et. al.}

\begin{abstract}
%금융학에서 chart pattern matching problem은 미리 디자인한 chart pattern과 chart가 match되었는 지 여부를 판별하는 문제로 광범위하게 연구되어 왔다. chart pattern matching은 그 자체로 chart pattern의 matching을 정의한다. 하지만 어떻게 더 나은 matching을 정의하고, 정의된 matching을 사용하여 매력적인 chart pattern을 search할 지에 대한 연구는 연구자들로부터 관심을 끌지 못하고 있다.
%이에 이 논문에서는 chart pattern에 대한 진보된 matching으로 neural network-based chart pattern을 제안하고 해당 matching에서 매력적인 chart pattern을 찾는 자동 알고리즘을 소개한다. neural network-based chart pattern은 non-linear변환을 포함하는 adaptive parameteric feature와 해당 feature space상에서 적용 가능한 template을 neural network으로 표현한다. 우리는 general chart pattern 탐색 문제를 최적화 문제로 공식화한다. 매력적인 neural network-based chart pattern을 search하기 위해 우리는 GPGPU 기반의 parallel computation으로 구현하여 빠른 평가를 보장하는 HyperNEAT를 활용한다. 제안한 framework을 사용하여 한국 주식 시장 데이터로 실험한 결과, 매력적인 chart patterns을 성공적으로 발견하였다. 끝으로 우리는 분석을 위해 차트 패턴을 시각화함으로써 주식 시장에 대한  이해를 높일 특성들을 관찰할 수 있었다.
%original:\\
%The chart pattern matching problem determines whether a chart matches to a pre-designed pattern. It has been extensively studied in the field of financial investment. However, works on how to find attractive chart patterns and which matching techniques are beneficial for chart pattern search have been relatively under-examined.\\
%modified:\\
A neural network-based chart pattern represents adaptive parametric features, including non-linear transformations, and a template that can be applied in the feature space. The search of neural network-based chart patterns has been unexplored despite its potential expressiveness.\\
In this paper, we formulate a general chart pattern search problem to enable cross-representational quantitative comparison of various search schemes. We suggest a HyperNEAT framework applying state-of-the-art deep neural network techniques to find attractive neural network-based chart patterns; These techniques enable a fast evaluation and search of robust patterns, as well as bringing a performance gain. The proposed framework successfully found attractive patterns on the Korean stock market. We compared newly found patterns with those found by different search schemes, showing the proposed approach has potential. 
 
%적용한 기술들은 빠른 평가를 가능하게 하고, robust pattern을 찾게하고, 성능상의 이점을 가져온다. 
% Finally, we visualized the found patterns for in-depth analysis, and thereby were able to observe characteristics that would enhance the understanding of the stock market.
\end{abstract}

%A neural network-based chart pattern represents adaptive parameteric features including non-linear transformations and a template that can be applied on the feature space. A templete-based chart pattern is considered as linear model of neural network-based chart pattern.



%\ccsdesc[300]{Computing methodologies~Matching}

% We no longer use \terms command


\maketitle

\section{Introduction} \label{sec1}
According to the famous EMH (efficient market hypothesis \cite{fama_efficient_1970}), all relevant information is fully reflected in asset prices and the assets are traded at fair values. It asserts that no one can consistently beat the market, so any effort to find such a strategy is worthless. As refutations of this, a considerable amount of empirical studies have been conducted \cite{banz_relationship_1981,keim_size_related_1983}. In particular, technical analysis focuses mostly on the historical record of price movement, assuming that the fluctuations of asset prices have useful information for investment.\\ 
This preliminary paper focuses on chart patterns in technical analysis.
% 의의 밝힘:
% chart pattern analysis의 의의는 그것이 trading strategy를 정립하는데 핵심적인 역할을 한다는 것이다. 가령, 복수의 수익성 패턴을 이용하여 long signal을 생성하거나 loss를 implying하는 패턴들로 short signal을 생성하는 간단한 strategy를 디자인할 수 있다. 다만 본 논문에서는 strategy 수준의 논의는 전개는 생략하고 chart pattern analysis에 집중한다.
The significance of the analysis is that patterns often play a key role in building a trading strategy.\footnote{Triggering actions using chart patterns is a common approach for trading strategy in practice. Since the study of strategy is beyond the scope of our research, it is not covered in this paper.} 
A templated-based chart pattern matching is an approach to determine whether a reference chart matches to a template \cite{bo_empirical_2005,cervello_royo_stock_2015,leigh_trading_2008,leigh_forecasting_2002,wang_stock_2007,wang_trading_2009,leigh_market_2002}. This approach has a strong limitation to detecting highly abstracted features. On the other hand, a rule-based approach enables one to design patterns with abstracted features, yet it has the disadvantage that experts must manually define syntactic elements and their structures \cite{bandara_complex_2015,fu_stock_2007,lee_pattern_2006,lo_foundations_2000,zapranis_novel_2012,zhang_real_2010,ha_genetic_2016}. Kamijo and Tanigawa \cite{kamijo_stock_1990} attempted to recognize the triangle chart pattern with a recurrent neural network as a precedent study of a neural network-based matching. Guo \textit{et al.} \cite{guo_automatically} used a rival penalized competitive learning (RPCL) neural network for clustering stock chart patterns. Unlike stock chart pattern analysis, the use of a neural network for the control chart pattern recognition has been actively studied in the field of statistical process control (SPC) \cite{addeh_control_2011,barghash_pattern_2004,leger_fault_1998,cheng_neural_1997}. \\
 % The control chart pattern and the stock chart pattern analysis have been studied from different domain, but they have many common characteristics.
Prior chart pattern studies have focused mostly on how to discriminate whether a chart matches to a chart pattern. They blindly used well-known patterns, such as the \textit{head-and-shoulder}, without sufficient consideration on how to design a profitable chart pattern. The well-known patterns are likely to have been generated by inefficient searches on very limited regions of the pattern space since they were designed by human intuition. Ha \textit{et al.} \cite{ha_genetic_2016} posed this problem and applied a genetic algorithm (GA) to find rule-based chart patterns automatically.\\
In this paper, we formulate a chart pattern search problem independent of how chart patterns are designed. 
% NNd의 사용 정당성을 밝힘
% chart pattern은 chart를 입력으로 받아 매치 여부를 출력하는 discriminant function 그 자체이다. (see Section 2.2). 한편 neural network은 이론적으로 universal approximation이 가능하여 optimal한 discriminant function을 approximate하는데 유리한 모델이다. 따라서 우리는 neural network으로 chart pattern을 표현하기에 적합하다고 판단한다. 하지만 chart pattern search는 전형적인 gradient based approach로 학습할 수 없는 문제이므로 우리는 neural network의 parameter 최적화 문제를 evolutionary computation을 기반한 방식으로 해결한다.
We propose a search framework using Hypercube-based NeuroEvolution of Augmented Topologies (HyperNEAT; \cite{gauci_autonomous_2010,stanley_hypercube_based_2009}) for template-based and neural network-based chart pattern searches, which have not been studied in the past. 
Neural networks are advantageous to approximate a function in theoretical reason (see Section~\ref{sec2.1}). On the other hand, a chart pattern is a discriminant function itself that takes a chart as input and outputs a match (see Section~\ref{sec2.2}). Therefore, a neural network is a suitable candidate to represent the chart pattern. Since typical gradient-based learning techniques are not applicable for the neural network-based chart pattern search, we solve it using the NeuroEvolution framework.
We implement the proposed framework applying the state-of-the-art deep neural network techniques: 
\begin{enumerate*}[label={\alph*)}, font={\bfseries}]
\item the use of GPGPU allows handling the tremendous cost of a stock chart pattern search;
\item a simple but powerful regularizer, dropout \cite{srivastava_dropout_2014}, helps find general patterns;
\item the use of the Rectified Linear Unit (ReLU; \cite{nair_rectified_2010}) brings performance gain, as well as enabling fast operation, compared to the conventional activation function, sigmoid. However, ReLU should be used with caution because the variance of activations of a layer can be amplified or attenuated over layers. Weight initialization tricks \cite{glorot_understanding_2010,he_delving_2015} for gradient-based learning provides insight for weight scaling of a phenotype network using ReLU, allowing an efficient exploration of deeper networks without suffering such a problem.
\end{enumerate*}
%The proposed framework for chart pattern search is not domain specific, but we limit the scope of this study to stock chart pattern analysis. 
We found some attractive stock chart patterns in the Korean market with the framework. 
% To demonstrate the superiority of our approach, we compared newly found patterns with the manually designed well-known patterns and the patterns found by the GA \cite{ha_genetic_2016}. 
The suggested framework found some profitable patterns, outperformed the other patterns.\\
% In addition, we visualize the found patterns by superimposing matched charts to analyze the patterns, thereby improving understanding of the pattern and the stock market.\\
The remainder of this paper is organized as follows. In Section~\ref{sec2}, we introduce various chart pattern matchings and formulate a chart pattern search problem. Section~\ref{sec3} describes the HyperNEAT framework for neural network-based chart pattern search and its implementation with state-of-the-art deep neural network techniques. Section~\ref{sec4} presents the preprocess and the experimental setup. In Section~\ref{sec5}, we demonstrate the performance of the proposed approach with the results and show the advantages of applied neural network techniques. Finally, we draw conclusions in the last section.

\section{Chart Pattern Search} \label{sec2}
Section~\ref{sec2.1} provides an overview of various chart pattern matching techniques. In Section~\ref{sec2.2}, we formulate the chart pattern search problem independent of a matching, which establishes a basis to compare patterns represented by different matching techniques.

\subsection{Chart Pattern Matching} \label{sec2.1}

In the applications of chart pattern analysis, a raw time series can be preprocessed to either a one-dimensional sequence or a two-dimensional image. Since either of them can be vectorized, a chart can be written in the form of $ \bm{x} = (x_{1}, x_{2}, ..., x_ {D}) ^ T$.\\
A template-based matching uses a template $ \bm{w} = (w_0, w_1, ..., w_D) ^ T$ to discriminate whether a chart matches by the sign of a linear combination with the chart. The discriminant function $f$ can be written as follows:
\begin{equation} \label{eq1}
    f(\bm{x})=\begin{cases}
        1, & \text{ if } w_0 + w_{1}x_{1} + ... + w_{D}x_{D} > 0 \\
        0, & \text{ otherwise}.
    \end{cases}
\end{equation}
We say that $ \bm{x} $ \textit{matches} to the corresponding chart pattern when the value of $ f (\bm{x}) $ is $ 1 $. 
A template-based pattern forms a linearly separated area in the input space; the matched input lies in this area.
This linear model not only limits the representation of the chart pattern, but also has a high bias.
Nonetheless, it has the advantage in the field of finance due to the robustness from low variances.\\
A feature-based matching, on the other hand, discriminates a chart through a linear combination of fixed nonlinear functions that extract features from input. %It extend expression of the pattern. 
%, of the form
%\begin{equation} \label{eq2}
%    f(x)=\begin{cases}
%        1 \text{ if } w_0 + \sum_{i=j}^{M-1}{w_{j}\varphi_{j}(x)} > 0, \\
%        0 \text{ otherwise.}
%    \end{cases}
%\end{equation}
%where $w = (w_0, ..., w_{M-1})^T$ and $\varphi = (\varphi_0, ..., \varphi_{M-1})$.
The nonlinear feature extraction function maps the input to the feature space. Thus, the feature extraction function is called the basis function because it generates the basis of the feature space. Feature-based matching has a limitation that feature extraction functions must be pre-defined by an expert. However, there are few known features in the chart pattern analysis that induce profitability.\\
An alternative is to introduce adaptive parameters to the basis function so that the parameter can be adapted while searching chart patterns. The neural network is one of the most widely studied models to handle adaptive parametric basis functions.
Theoretically, a two-layer neural network containing a sufficient finite number of hidden nodes can approximate a continuous function defined on a compact domain \cite{cybenko_approximation_1989,hornik_approximation_1991}, therefore, the neural network-based chart pattern representation using adaptive parametric feature have potential expressiveness.
We use a neural network to represent a chart pattern.
Furthermore, a template-based chart pattern can be represented by a neural network with no hidden neurons.\\
%We review the rule-based chart pattern analysis for later experimental comparison(ref). 
In rule-based chart pattern analysis, a rule indicates the order relation of extrema extracted from a smoothen time series. A rule-based chart pattern consists of a set of rules. If the chart satisfies all the rules of the rule-based chart pattern, it is said to match. Most manually designed well-known patterns are rule-based. The \textit{head-and-shoulder}, for example, is expressed in the order relations of the extrema, the head, the two shoulders, and the two necks. 
% A rule-based chart pattern has a discriminant function that outputs 0 or 1 depending on whether a chart satisfies the rules of the pattern.

\subsection{Problem Formulation} \label{sec2.2}

%We clarify the notions regarding the chart pattern analysis. 
A \textit{chart pattern} itself can be considered to be a function, which discriminates whether a chart matches to the chart pattern. The \textit{chart pattern matching} describes how a discriminant function is represented. The feasible functions from a chart pattern matching span a discriminant function space. The \textit{chart pattern search} is to find an attractive chart pattern on the chart pattern space defined by a matching. Equivalently, it is a problem of finding an attractive discriminant function in the function space.\\
We define the profitability of a pattern by an expected value of log returns\footnote{The losses and profits are likely to be distributed fairly evenly. In practice, losses lie in the interval of $(-1, 0)$ and profits in the interval of $(0, \infty)$, which makes their distribution skewed severely to the profit side. Thus, we use the log returns that are more sensitive to losses.} after $ k $ days for all charts matched to the pattern. We define a chart pattern search as a problem of finding a discriminant function $f$ of a pattern such that it maximizes the profitability of the chart pattern. If the number of matches of a chart pattern is too small, the pattern is not general enough as well as it is hard to be used in the field. Thus, a penalty related to the number of matches is applied. The penalty function $g$ is defined as follows:
%The inputs matched to the chart pattern form an area on the inputs space. 
%
\begin{equation} \label{eq3}
    g(f,\bm{X})= \exp\Big(- \frac{6}{\alpha}\sum_{\bm{x}\in \bm{X}}{f(\bm{x})}\Big),\\
\end{equation}
where $\bm{X}$ is the set of all given charts, $\sum_{\bm{x}\in \bm{X}}{f(\bm{x})}$ is the number of matches, and $\alpha$ is a hyperparameter that determines the degree of penalty. 
When the number of matches is greater than $\alpha$, the panalty value is small enough to be ignored.
% The more the number of matches, the closer to zero  is increased, the penalty become close to zero.
For a given set of charts, the chart pattern search problem can be represented by the following optimization problem:
\begin{equation} \label{eq4}
    \resizebox{0.98\linewidth}{!}{%
    $Chart\_Pattern\_Search(k,\bm{X}) =  \underset{f}{\arg\max}\Big( \frac{\sum_{\bm{x}\in \bm{X}}r(\bm{x},k)f(\bm{x})}{\sum_{\bm{x}\in \bm{X}}f(\bm{x})}g(f,\bm{X})  \Big),$
    }
\end{equation}
where $r(\bm{x}, k)$ is the $k$-day log return after the last day of $\bm{x}$.  The chart pattern search problem is defined independently of what representation is used.

\section{HyperNEAT for chart pattern search} \label{sec3}

%\begin{table}[]
    %\centering
    %\caption{Neural network configuration}
    %\label{my-label}
    %\begin{tabular}{|c|}
        %\hline
        %\textbf{chart input}  \\ \hline
        %32x2                  \\ \hline
        %\textbf{hidden layer} \\ \hline
        %16x12                 \\ \hline
        %16x6 \\ \hline
        %8x6\\ \hline
        %4x6\\ \hline
        %4x3\\ \hline
        %\textbf{output} \\ \hline
        %1\\ \hline
    %\end{tabular}
%\end{table}

\begin{figure}
\includegraphics[width=1.0\linewidth]{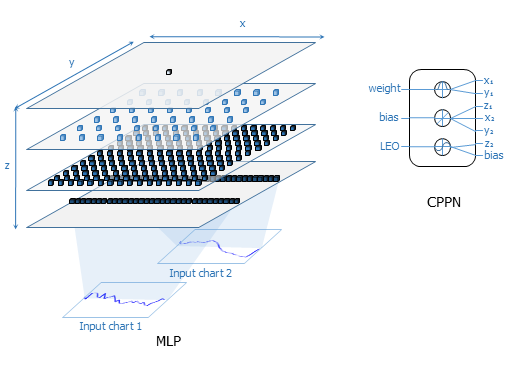}
\caption{proposed HyperNEAT}
\label{fig:hyperneat}
\end{figure}

The discriminant function of a chart pattern can be expressed as a neural network. This section describes how to generate neural networks to represent a chart pattern. HyperNEAT is designed to encode neural connectivity patterns with regularities such as symmetry, repetition, and deformation of neural networks. In particular, these patterns arranged on the substrate are geometrically interpreted and indirectly encoded by a Compositional Pattern Producing Network (CPPN) \cite{stanley_compositional_2007,stanley2006exploiting}. A CPPN receives a pair of coordinates of two neurons in a phenotype neural network as inputs and outputs their connection weight, or takes one neuron's coordinate with a zero-filled null coordinate to determine the bias through a separate output node. In addition, a CPPN recieves constant one as bias of genotype network in our implementation. This approach permits a relatively small genotype network to effectively describe neural connectivity patterns of a much larger phenotype network. The extension of HyperNEAT, called HyperNEAT with Link Expression Output (HyperNEAT-LEO), was introduced to limit connectivity with a bias towards modularity \cite{verbancsics_constraining_2011}. This creates a separate output called LEO as well as the weight and bias output of the genotype CPPN. LEO determines the expression of the link in the phenotype network. 
% Connective CPPN-NEATs allow evolution of connection patterns.
HyperNEAT has shown to be effective for a variety of problems \cite{clune_evolving_2009,dambrosio_evolving_2010,gauci_autonomous_2010,verbancsics_evolving_2010}. Nevertheless, to the best of our knowledge, no research exists to find a stock chart pattern with HyperNEAT. We present the HyperNEAT framework for the stock chart pattern search. We use the HyperNEAT-LEO version except for a few modifications. We adopt a multi-layer perceptron (MLP) architecture for a discriminant function of the chart pattern. After preprocessing, a chart input is represented by a $32\times2$ real-value matrix. The substrate configuration of MLP for a template-based chart pattern is composed of a $32\times2$ input layer and a $1\times1$ output layer, and the one for a neural network-based chart pattern is composed of $16\times12$ and $8\times6$ hidden layers between the input and output layers. To encode the phenotype indirectly, the geometric layout of MLP needs to be determined in advance. Figure~\ref{fig:hyperneat} shows the actual configuration of the neural network-based chart pattern. 
The $x$, $y$, and $z$ denote the time axis of input, the input channel axis, and the layer axis, respectively. 
% Let the time axis of input be $x$, the input channel axis $y$, and the layer axis $z$. 
The layers are arranged equally dividing the interval $[-1.0,1.0]$ and the endpoints. The nodes are placed in the coordinates equally dividing the interval $[-1.0,1.0]$ for each of $x$ and $y$ axes.  When the value of the LEO output is not greater than zero, the connection is not expressed, thereby the weight of connection is set to zero. The MLP receives a chart and outputs 0 or 1 to determine chart matching. After obtaining candidate matches of a pattern, it is evaluated based on the expected profitability.\\
%??? one based on the evaluation of the expected profitability.
% the expected profitability can be calculated to evaluate one organism.\\
We adopt the state-of-the-art deep neural network technique for computationally efficient and effective implementation of a phenotype neural network of HyperNEAT.
\paragraph{GPGPU for fast evaluation}
We use about one million input charts of the training set for the experiments (see Section~\ref{sec4}). Therefore, a phenotype network should evaluate over one million charts to evaluate one organism. Because these operations require significant computational resources, efficient computational design will determine the possibility of practical experiment.\\
In recent deep neural network studies, the use of GPGPU for deep and complex neural networks using considerable amounts of data has been applied to various problems and shows excellent performance \cite{krizhevsky_imagenet_2012,he_deep_2015}. We use GPGPU for fast evaluation of phenotype networks. The feedforward operation of the MLP structure is composed of simple matrix computation such as multiplication and summation so that the gain of GPGPU can be fully exploited. TensorFlow is a library for numerical computation using stateful data flow graphs \cite{tensorflow2015_whitepaper}. It allows flexible CPU or GPU selection for numerical computation without consideration of low-level implementation. We use TensorFlow to take advantage of GPGPU from fast phenotype network operations.
\paragraph{Dropout}
Dropout \cite{srivastava_dropout_2014} is a regularization technique to prevent coadaptation of feature detectors by randomly and temporarily omitting nodes of a neural network. We use it to prevent a neural network from overfitting to training set. The probability of retaining a neuron is set to 0.8 for evaluation using training set.
\paragraph{Rectified Linear Unit and weight scaling}
ReLU \cite{nair_rectified_2010} is an activation function defined by $h(a) = max(0, a)$ where $a$ is a preactivation of neuron.
It is known to be a practical solution of the vanishing or exploding gradient problems of gradient-based learning, but it also has the effect of reducing the computational cost compared to the traditional activation function, i.e., sigmoid. We use ReLU as an activation function in our experiments, which is more cost-effective and provides a performance improvement.
Because the ReLU propagates the variance of positive valued activation to the next layer, the variance can be amplified or attenuated over layers.
Careful weight initialization techniques \cite{glorot_understanding_2010,he_delving_2015} for deep learning were introduced to prevent this problem, so it became a key factor for designing deep architectures. They carefully scale randomly selected weights so that the variances of activation over layer are kept constant. We scale the CPPN's weight outputs from the same insight. We refer to He \textit{et al.}'s work \cite{he_delving_2015} and scale the phenotype neuron's weight to $\sqrt{(2.0/N_{in})}$ times where $N_{in}$ is the number of connected neurons from the preceding layer.\\
% This allows one to compute larger networks or to handle more extensive evolutionary computations with the same computational cost.

HyperNEAT evolves connective CPPN through NeuroEvolution of Augmenting Topologies (NEAT) \cite{stanley_evolving_2002}. NEAT is a genetic algorithm allowing the evolution of neural networks. A neural network has a permutation problem which makes difficult to define crossover, but it overcomes the problem by using historical markers. Beginning with a population of small and simple neural networks, this algorithm makes them increasingly complex. It exploits and explores with evolutionary tinkerings, such as adding new nodes and connections to a neural network. In addition, to preserve the topological innovation, speciation is applied by sharing fitness among similar individuals.
We set the population size to 1000; it is carried out until reaching the 200th generation. A decay factor of 0.999 is introduced to decrease the mutation rates gradually over generations. 
%The mutation rates are decreased by the factor of 0.999 every generation. 
In addition, when the compatibility threshold is fixed, the population speciate too much. Since it cannot take advantage of niching, we raised the threshold to the rate of 1.001 per generation. For the same reason, if the number of species exceeded 100, the degree of speciation is adjusted by raising the threshold by 1.1 times. We do not allow for a genotype CPPN to contain any recursive connection since charts are assumed to be independent and identically distributed in chart pattern analysis.\\
% CPPN은 constant one을 bias 입력으로 받는다. 

\section{Experimental Setup} \label{sec4}
This section describes preprocessing and experiments for the chart pattern search problem.

\subsection{Preprocessing} \label{sec4.1}
\begin{figure}
\includegraphics[width=1.0\linewidth]{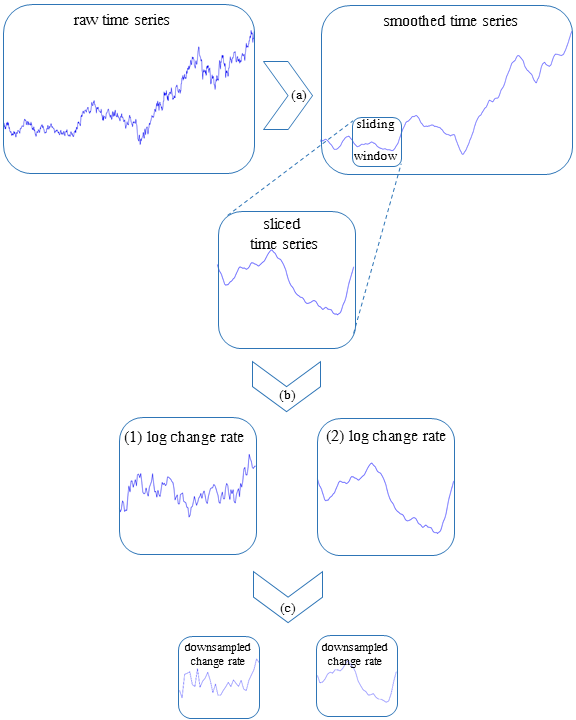}
\caption{Preprocessing}
\label{fig:preprocess}
\end{figure}

Figure~\ref{fig:preprocess} shows the preprocessing for the chart pattern analysis used in this study.
In the chart pattern analysis, smoothing techniques such as moving average, linear Kalman filter, and kernel regression are used \cite{bandara_complex_2015} because raw financial time series data tend to have unstable fluctuations in short term. We smoothed the time series using a 24-day moving average as shown in Figure~\ref{fig:preprocess}a. 
The sliding window is shifted by one day in each time series and emits a sliced time series. If a sliding window of size $s$ is used in $l$-length time series, then $l - s + 1$ sliced time series are obtained. We used 128 days for window size. Each sliced time series was processed into two channels of chart as shown in Figure~\ref{fig:preprocess}b: 
\begin{enumerate*}[ font={\bfseries}]
\item the logarithmic rate of daily price change;
\item the logarithmic rate of price change of a day to the last day.
\end{enumerate*}
We took logarithm to resolve the skewedness. 
In addition, the preprocessed data were downsampled by a factor of 4 to reduce the size of the preprocessed data as shown in Figure~\ref{fig:preprocess}c.
Finally, we reduced the magnitude of the second channel by a factor of $1/32$ to match the one of the first channel.
Apart from this, we computed the $k$-day return rate of the day after the last day of the preprocessed data.
In a template-based chart pattern analysis, two-dimensional chart images are popularly used as input. However, the height of the two-dimensional chart is determined in advance, so the magnitude of the price or the change rate is adjusted according to the height. We use the one-dimensional real-valued input to maintain magnitude information.

\subsection{Experimental Setup and Parameters} \label{sec4.2}

We seek attractive patterns that operate on the stock market, rather than on a particular stock or indicator. Therefore, we used the daily closing prices of all stocks ever-listed in the Korean stock market from January 2012 to December 2016. We divided the dataset into three groups:
\begin{itemize}
    \item training set: January 2012 through December 2014,
    \item validation set: January 2015 through December 2015,
    \item test set: January 2016 through December 2016.
\end{itemize}
The numbers of samples were 1018045, 294586, and 292141 for training, validation, and test sets, respectively.
The training set is used for optimizing chart patterns, the validation set for model selection, and the test set for assessment of the approach.
We refined the datasets by excluding both preferred and fund stocks on the stock market. Unlike the US stock market, the Korean stock market has both upper and lower limits of daily price movement. If the price of the item reaches the daily upper limit, it is extremely hard to buy the item. Therefore, it is assumed that the chart is not matched in that case.\\
We experimented the $Chart\_Pattern\_Search(k, \bm{X})$ of genetic frameworks for the various representations for quantitative comparisons. 
Template-based and neural network-based chart pattern searches were conducted with the proposed HyperNEAT framework.
In addition, we re-implemented the GA \cite{ha_genetic_2016} to search rule-based chart patterns for the given datasets.
The hyperparameter $\alpha$ was set to 100000. We tuned solutions for each objective function by varying the parameter $k$ in three ways: 20, 50, and 100. Our implementation of Python and TensorFlow took about 33 hours to run on an i7-4770 CPU @ 3.40GHz machine with a GeForce GTX 980 GPU card.

\begin{table*}[h!]
    \begin{center}
        \caption{Comparative results ($\times10^{-2}$)}
        \label{table1}
        \resizebox{\linewidth}{!}{
            \begin{filecontents*}{result.csv}
Approach,Pattern name,train20,valid20,test20,train50,valid50,test50,train100,valid100,test100
,doubleB,-0.4135,0.1745,-0.713,0.0299,3.001,-0.1425,0.8759,6.3533,1.3904
,doubleT,-0.1833,0.0892,0.0562,0.1842,1.5489,0.6741,1.4297,4.4436,\textbf{2.5205}
,triangleA,0.0077,-0.0138,-0.0361,0.0007,0.0201,0.0162,0.0222,0.0442,0.0502
,triangleD,0.0142,-0.0079,-0.0296,0.0418,0.0937,-0.0818,0.0287,0.1796,-0.0414
,triangleSB,0.0168,0.0154,0.0134,0.0411,0.0637,0.0051,0.0385,0.1354,0.1185
,triangleST,-0.0104,0.002,-0.0643,0.0284,0.0684,-0.019,0.0434,0.093,-0.0056
,tripleB,-0.1011,0.0444,-0.1613,-0.0113,0.8624,0.0138,0.2487,2.3948,0.903
,tripleT,-0.0262,-0.0075,-0.0375,-0.0349,0.1079,0.0674,0.1352,0.2817,-0.0999
,threeFP,-0.0139,\textbf{0.5688},\textbf{0.1183},\textbf{1.5674},\textbf{3.6929},\textbf{0.8612},\textbf{3.9923},\textbf{8.997},2.439
,threeRV,-0.2771,0.2085,-0.8037,-0.1859,2.2642,-0.2382,0.2884,4.1644,1.2385
,HnSB,-0.0028,-0.0013,-0.0065,-0.0018,0.0276,-0.0203,0.0614,0.0612,0.0229
Manually designed,HnST,-0.0042,0.0102,-0.0249,0.0165,-0.0077,-0.0384,0.0248,0.1369,-0.0625
well-known,broadenB,-0.0068,0.0008,-0.0135,-0.0111,0.0142,-0.01,-0.0029,0.0006,-0.0071
stock chart patterns,broadenT,0.0055,0.0022,-0.0059,0.0102,-0.0043,-0.0025,0.012,-0.0132,0.0044
,broadenFA,0.0023,-0.0011,0.0031,0.0197,0.0233,-0.0069,0.1019,0.0968,0.0554
,broadenFD,-0.0051,0.005,-0.0117,-0.0073,0.0192,-0.0046,-0.0003,0.0567,-0.0081
,broadenWA,-0.0665,-0.0397,-0.0978,-0.0362,0.0819,-0.0696,0.0459,0.5141,0.1591
,broadenWD,\textbf{0.0358},0.1306,0.0189,0.2502,0.4267,0.1748,0.5445,0.9414,0.7562
,rectB,-0.002,-0.0159,-0.0071,-0.0043,0.0016,-0.0252,-0.0042,0.0034,-0.0206
,rectT,-0.0024,-0.0172,-0.0024,-0.0091,0.0098,-0.0116,-0.0063,0.0067,-0.0142
,BnRB,0.0313,0.1773,0.0135,0.2508,0.6398,0.2824,0.6682,1.1413,0.9724
,BnRT,-0.0892,-0.0475,-0.1095,-0.0415,0.0705,-0.0556,0.079,0.2984,0.225
,DHnSB,-0.0011,-0.0006,0.0014,-0.0071,-0.006,-0.0175,-0.0015,0,-0.0223
,DHnST,0.0006,0.0033,0.0011,0.0013,0.013,0,0.0004,0.018,0
,diaB,-0.0002,0.0042,-0.0046,0.0264,0.0145,0.0028,0.0146,0.0086,0.0153
,diaT,-0.0105,0.0123,0.0022,-0.0021,0.0236,0.0121,-0.006,0.0167,0.0195
,HnDSB,-0.0012,0,0,-0.0034,0,0,-0.0001,0,0
,rp\_20\_50\cite{ha_genetic_2016},-0.1657,0.6548,-0.8153,1.4888,6.4532,\textbf{0.0863},3.2787,12.2918,1.9913
Rule-based,rp\_100\cite{ha_genetic_2016},-0.1127,0.5155,-0.5296,0.3814,2.2943,-0.3843,1.0427,4.9562,0.7167
stock chart patterns,rp\_example\cite{ha_genetic_2016},-0.2185,0.4883,-0.8412,0.9535,4.7315,-0.6867,2.3514,9.7325,0.9351
found by GA,rp2\_20,0.2112,\textbf{1.4374},-0.3566,3.0874,6.1921,-0.4915,5.2878,9.9223,0.5566
,rp2\_50,\textbf{0.3646},1.3335,\textbf{-0.2888},\textbf{3.2314},\textbf{6.5954},-0.1334,\textbf{6.5452},15.4184,2.3219
,rp2\_100,0.3277,0.0994,-0.4351,1.5327,4.6352,-0.0762,5.8328,\textbf{16.3815},\textbf{2.6038}
Template-based,tp\_20,9.9308,4.151,2.2809,6.8905,2.6877,0.4456,4.7645,4.1592,-4.5077
stock chart patterns,tp\_50,\textbf{14.8681},\textbf{8.0650},\textbf{6.1008},\textbf{11.9267},\textbf{7.2954},5.338,\textbf{9.302},9.8671,-0.0063
found by HyperNEAT,tp\_100,-1.7451,0.8408,2.9105,0.9945,5.0631,\textbf{7.0560},4.4891,\textbf{15.8227},\textbf{4.3086}
Neural network-based,nnp\_20,\textbf{14.9717},\textbf{8.0733},$[$\textbf{6.1778}$]$,\textbf{12.0797},\textbf{7.0082},5.2648,\textbf{9.5376},9.5248,-0.1256
stock chart patterns,nnp\_50,0.3281,1.2617,2.9870,3.5819,5.6478,7.1354,5.8694,14.5219,4.8035
found by HyperNEAT,nnp\_100,-0.6693,1.2977,3.002,2.0989,5.2279,$[$\textbf{7.3442}$]$,5.5765,\textbf{15.4207},$[$\textbf{5.5817}$]$
\end{filecontents*}
\pgfplotstabletypeset[
                precision = 4,
                col sep = comma,
                fixed relative,
                string replace*={_}{\textsubscript},
                every head row/.style={
                    before row={%
                        \toprule
                        &\multicolumn{1}{c}{} & \multicolumn{3}{c}{$k=20$} & \multicolumn{3}{c}{$k=50$} & \multicolumn{3}{c}{$k=100$}\\
                    },
                    after row=\midrule,
                },
                % every row no 0/.style={after row=\midrule},
                every row no 26/.style={after row=\midrule},
                every row no 32/.style={after row=\midrule},
                every row no 35/.style={after row=\midrule},
                every last row/.style={
                    after row=\bottomrule
                },
                columns/train20/.style ={column name=Training},
                columns/valid20/.style ={column name=Validation},
                columns/test20/.style={column name=Test},
                columns/train50/.style ={column name=Training},
                columns/valid50/.style ={column name=Validation},
                columns/test50/.style={column name=Test},
                columns/train100/.style ={column name=Training},
                columns/valid100/.style ={column name=Validation},
                columns/test100/.style={column name=Test},
                string type
                %every head row/.style={before row=\toprule,after row=\midrule},
                %every last row/.style={after row=\bottomrule},
                %display columns/0/.style={string type,column name={}}
            ]
            {result.csv} % filename/path to file
        }
    \end{center}
\end{table*}

\section{Experimental Results} \label{sec5}
This section provides a comparative analysis of the patterns found and the benefits of the key technologies applied for a neural network.

\begin{figure*}[t!]
    \centering
    \begin{subfigure}[b]{0.33\textwidth}
        \includegraphics[width=\textwidth]{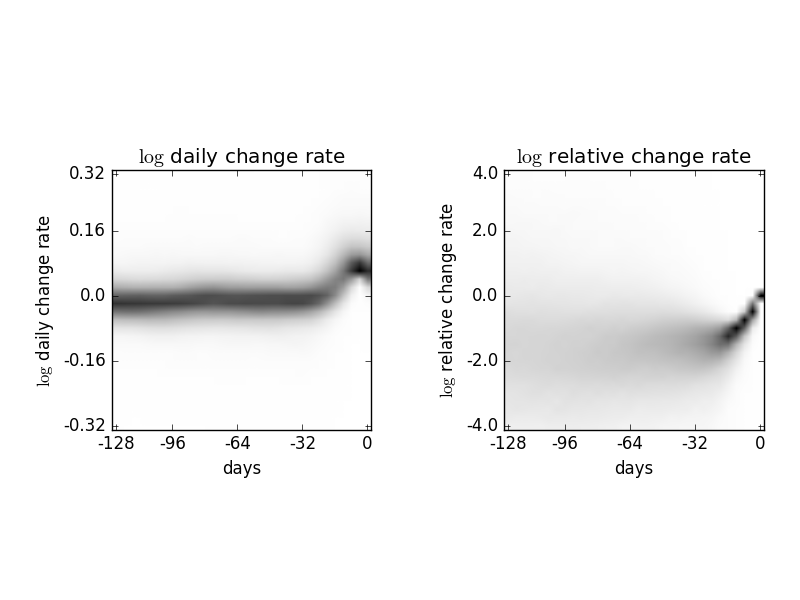}
        \caption{tp\_20}
        \label{fig:tp0}
    \end{subfigure}
    ~
    \begin{subfigure}[b]{0.33\textwidth}
        \includegraphics[width=\textwidth]{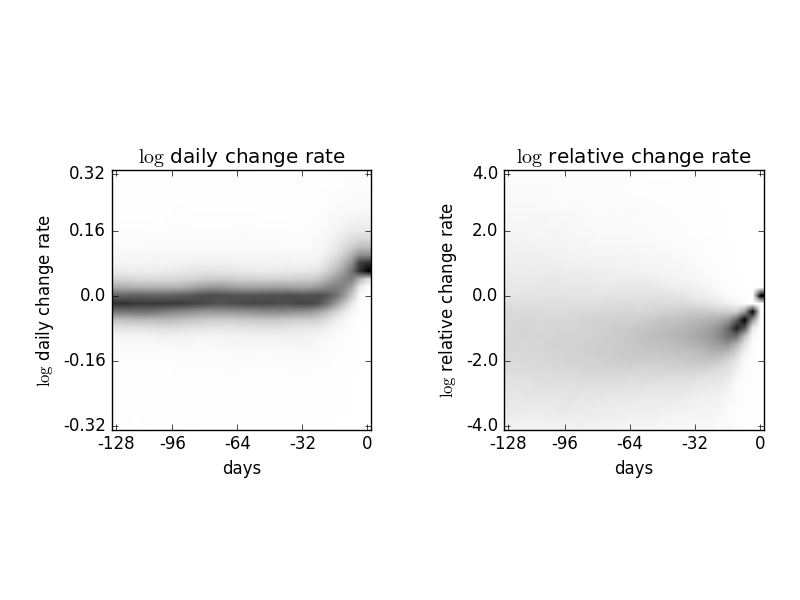}
        \caption{tp\_50}
        \label{fig:tp1}
    \end{subfigure}
    ~
    \begin{subfigure}[b]{0.33\textwidth}
        \includegraphics[width=\textwidth]{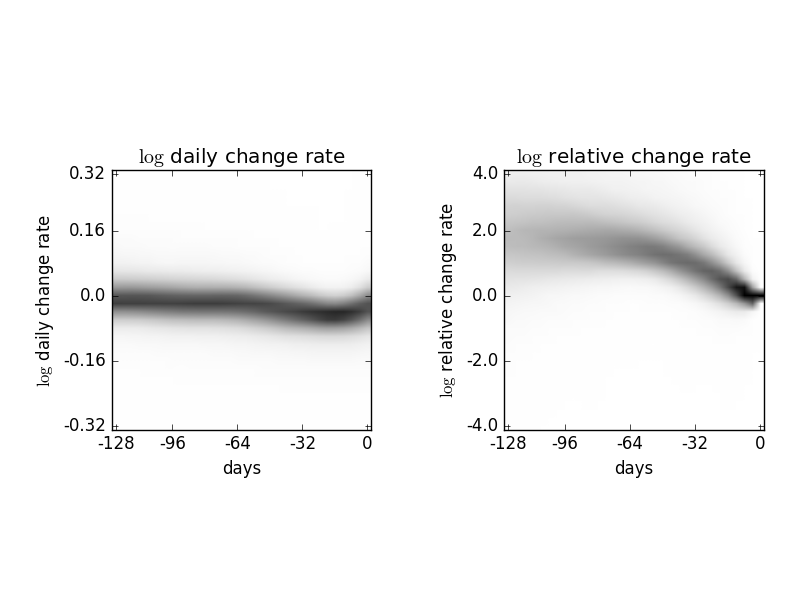}
        \caption{tp\_100}
        \label{fig:tp2}
    \end{subfigure}
    \begin{subfigure}[b]{0.33\textwidth}
        \includegraphics[width=\textwidth]{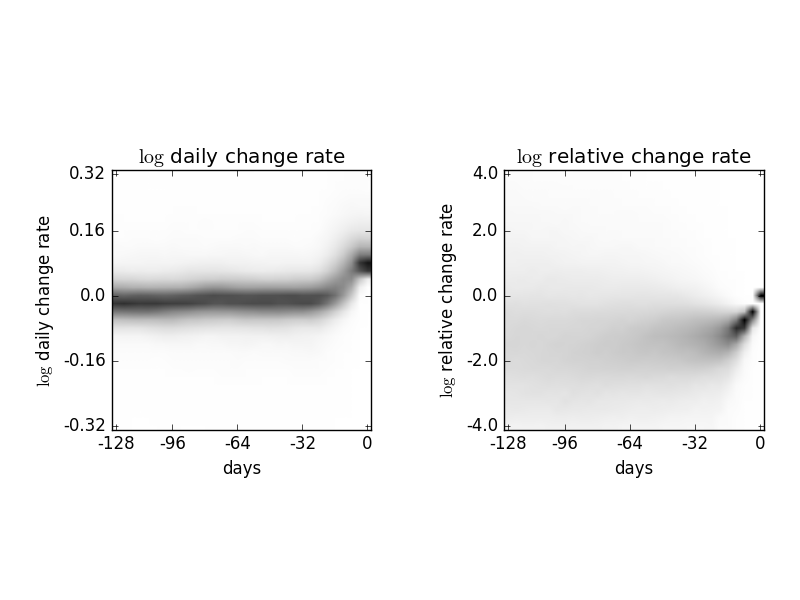}
        \caption{nnp\_20}
        \label{fig:np0}
    \end{subfigure}
    ~
    \begin{subfigure}[b]{0.33\textwidth}
        \includegraphics[width=\textwidth]{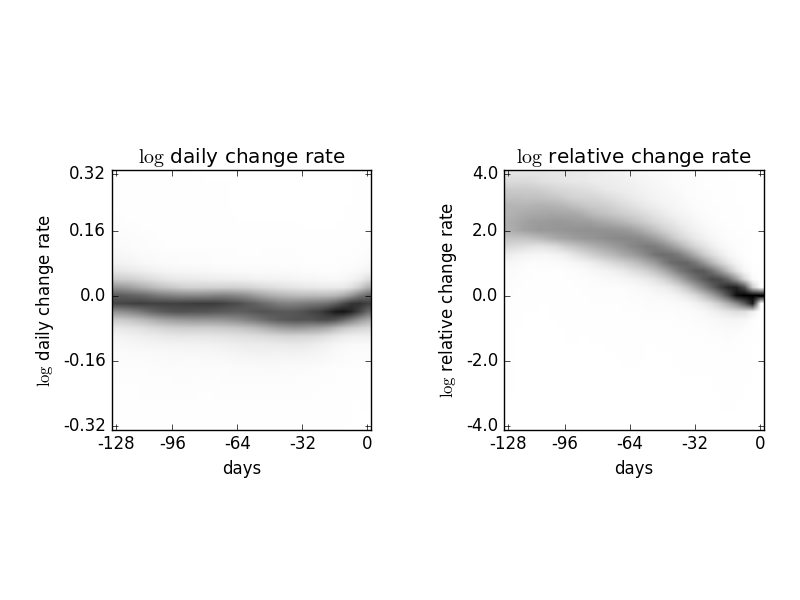}
        \caption{nnp\_50}
        \label{fig:np1}
    \end{subfigure}
    ~
    \begin{subfigure}[b]{0.33\textwidth}
        \includegraphics[width=\textwidth]{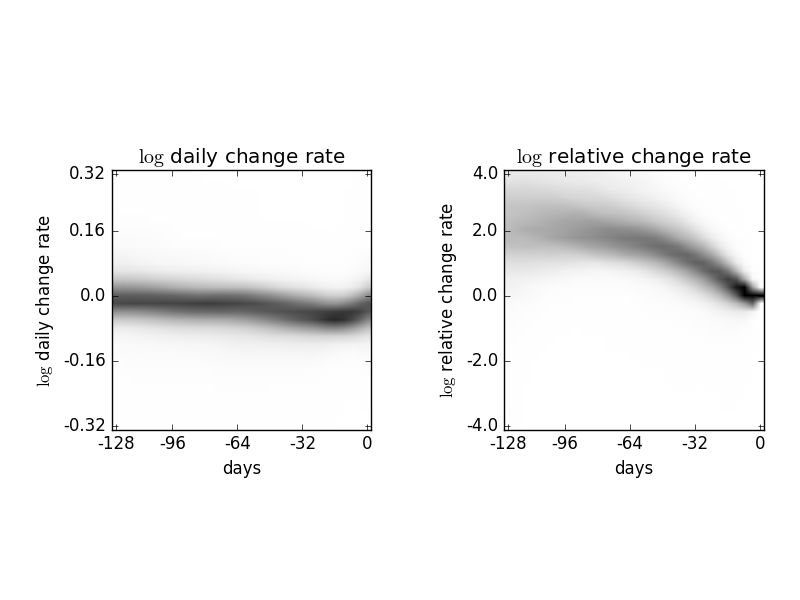}
        \caption{nnp\_100}
        \label{fig:np2}
    \end{subfigure}
    \caption{Examples of patterns found}
    \label{fig:patterns}
\end{figure*}

\subsection{The Chart Patterns Found} \label{sec5.1}

We conducted a full-scale examination of well-known patterns\footnote{Manually designed well-known 26 patterns are the following: \textit{double-bottom, double-top, triangle-ascending, triangle-descending, triangle-symmetric-bottom, triangle-symmetric-top, triple-bottom, triple-top, three-falling-peak, three-rising-valley, head-and-shoulder-bottom, head-and-shoulder-top, broaden-bottom, broaden-top, broaden-formation-ascending, broaden-formation-descending, broaden-wedge-ascending, broaden-wedge-descending, rectangle-bottom, rectangle-top, bump-and-run-reversal-bottom, bump-and-run-reversal-top, double-head-and-shoulder-bottom, double-head-and-shoulder-top, diamond-bottom, diamond-top, head-and-double-shoulder-bottom}, and \textit{head-and-double-shoulder-top}.} and the found patterns using genetic frameworks for the various matchings. All the well-known patterns were provided in a rule-based representation \cite{ha_genetic_2016}. We used GA \cite{ha_genetic_2016} to find the profitable patterns for rule-based representation. The patterns found in the previous study are \textit{rp\_20\_50, rp\_100}, and \textit{rp\_example}, and the patterns found in the updated dataset in this paper were \textit{rp2\_20, rp2\_50} and \textit{rp2\_100}. We used the HyperNEAT framework proposed in this paper to find attractive template-based and neural network-based chart patterns. The names of the patterns found for each $k$ are: \textit{tp\_20, tp\_50}, and \textit{tp\_100} for template-based representation and \textit{nnp\_20, nnp\_50}, and \textit{nnp\_100} for neural network-based representation.\\
Table~\ref{table1} shows the performance of each pattern. The columns represent each dataset for each $k$ value, and the rows represent patterns. If no chart matches to a pattern, the fitness of the pattern is marked as zero. 
%The figures are enlarged 100 times for readability. 
% We show the performance of the pattern matching all the charts at the top of the table so that we can compare with the market average. 
Bold figures indicate the best column values for a given $k$ for each cluster of patterns. 
The values enclosed in square brackets are the best patterns out of all the patterns for a given $k$ in terms of test sets. When $k$ is 20, \textit{nnp\_20} was the best, and when $k$ is 50 or 100, \textit{nnp\_100} was the best. Overall, we observed that neural network-based chart patterns outperformed the other clusters. In particular, they showed consistently high profitability with an acceptable variance over all datasets. Templated-based patterns followed neural network-based patterns in a comparable range. For the rule-based patterns, it seems to be more difficult to find a general pattern with $k = 20$, the lowest.\\
We visualized the chart patterns found for a closer look at. Figure~\ref{fig:patterns} shows superimposed charts that match to the each found patterns to see the actual images as well as the characteristics of the patterns. Each of the two images of an input channel is separately piled up. The logarithmic rate of daily price change is on the left side and the logarithmic rate of price change of a day to the last day is on the right side. The more overlay, the darker colored. The horizontal axis indicates the day and the vertical axis indicates the magnitude.\\
As shown in Figure~\ref{fig:patterns}, we found similar patterns in template-based and neural network-based chart patterns, which can be roughly classified into two classes. One is a soaring pattern represented by \textit{tp\_20} (Figure ~\ref{fig:tp0}), \textit{tp\_50} (Figure~\ref{fig:tp1}), and \textit{nnp\_20} (Figure~\ref{fig:np0}), and the other is a falling pattern represented by \textit{tp\_100} (Figure~\ref{fig:tp2}), \textit{nnp\_50} (Figure~\ref{fig:np1}), and \textit{nnp\_100} (Figure~\ref{fig:np2}). 
The representative patterns found were rather simple. We suspect that many attractive patterns are hidden in the pattern space, considering our modelling is preliminary.
%The soaring pattern has little change until about $-20$th day, while the price gradually increases after then. This upward trend seems to be maintained for a relatively short period of time so that the 20-day and 50-day returns are high. On the other hand, the falling pattern continues to decline from the beginning, and the declining trend is accelerating after around $-60$th day. From about $-15$th day, the downtrend gradually eased. 
%%On the 0th day, the downtrend seems to end. 
%In this case, the downward trend seems to be reversed and the price gradually increased, indicating that profitability is high after about 50 and 100 days. High-yielding patterns in the market were simple rather than complex, and could be explained without violating human intuition.\\
%Figure~\ref{fig:milestone}  shows the history of the best patterns of the population for \textit{nnp\_50} by the same technique as the previous figure. The axes of each chart are the same as Figure~\ref{fig:patterns}, and the generation and fitness of the pattern are captioned on the top of the pattern. In earlier generations, there is an interval where the drop decreases at the end of the chart, while the interval disappears from the 22nd generation. In addition, we observed that the patterns could escape without staying at the peak formed at the 15th generation and move to the 85th generation peak. 
% The proposed HyperNEAT framework gradually develops patterns to find near optimal solutions. 

\subsection{Advantages of state-of-the-art neural network techniques} \label{sec5.2}
We examined the benefits of each state-of-the-art deep neural network technique.

\paragraph{GPGPU}
The computational time reduction for MLP is important given that most of the computational costs arise from the evaluation of phenotype networks. We were able to take advantage of GPGPU for evaluation of MLP. We compared the efficiency of GPGPU with CPU computation on the same TensorFlow implementation of evaluation of \textit{nnp\_50} 100 times for each. It takes about 587 seconds to evaluate by CPU operation, while about 21 seconds by the GPU operation. This results in a computational gain of about 27 times. It allows us to search chart patterns within tolerable amount of time.
% 다시 재볼 것
\paragraph{Dropout}
\begin{figure}[t!]
\centering
    \includegraphics[width=0.5\textwidth]{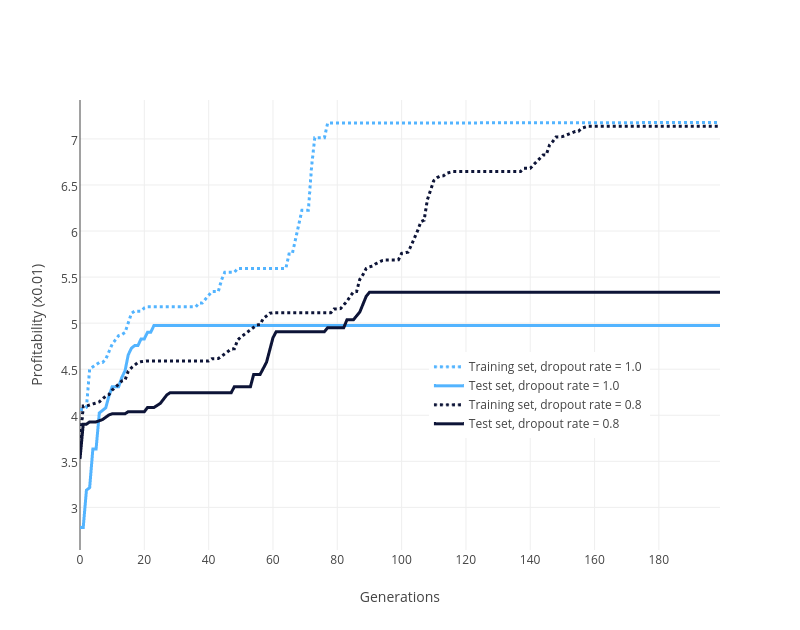}
    \caption{Average performance according to dropout rate}\label{fig:dropout}
\end{figure}

As shown in Figure~\ref{fig:dropout}, we compared the average performance difference according to whether the dropout was applied or not on 5 runs of neural network-based chart pattern search when $k$ is 100. The darker color is used when dropout is applied, and the training set is represented by a dotted line to distinguish from test set. In the case without dropout (dropout rate = 1.0), the average performance in the training converged relatively faster and the final performance is slightly better than the other case. However, the average performance of the test set was not improved after about 20th generation, resulting in overfitting to training set. On the other hand, in the case with dropout (dropout rate = 0.8), the average performance of training set was improved relatively slowly until about 160th generation. Furthermore, the average performance of test set was also improved slowly, so it outperformed the other case from around 80th generations. Therefore, dropout served as a significant regularizer to prevent to overfit to training set.

\paragraph{ReLU and weight scaling}
The use of ReLU as an activation function of the phenotype network has a slight benefit in computation time compared to the sigmoid network. 
%In detail, it takes about 21.371 seconds to evaluate the \textit{nnp\_50} 100 times using the ReLU network, but about 22.596 seconds using the sigmoid network.
However, without properly scaled weights, it is difficult to find attractive patterns using the ReLU network due to the occurrence of amplification or attenuation of the variance of activations over layers. We tested each activation functions with deeper neural networks to observe the amplication or attenuation; The substrate configuration of MLP was composed of a $32\times 2$ input layer, $16\times 12$, $16\times 6$, $8\times 6$, $4\times 6$, and $4\times 3$ hidden layers and $1\times1$ output layer. The population size was set to 100 and ran 5 times for each.
In the early generation of the ReLU networks, the average variance of the weights was about 0.096 over the whole network and the variance of activation was amplified to about $6.2\times10^7$ times from the input to preactivation of output. The CPPN could not successfully suppress this amplification over 200 generations, so it did not significantly improve the performance from the beginning.
By introducing a scaling factor, following He \textit{et al.}'s suggestion \cite{he_delving_2015}, we scaled the weights and managed variances. It enabled us to bound the amplification in a few thousand times, which is a comparable variance level of the sigmoid networks. The best fitness of the ReLU network was 0.051.
On the other hand, the amplification problem did not occur when sigmoid was used in phenotype network, so the performance did not deteriorate.
Furthermore, we could not observe the performance difference depending on the weight scaling for the sigmoid networks. The best fitness of the sigmoid network was 0.049.
The use of ReLU with careful weight scaling provided better record with reduction of the computation time by about 5.421\%.
\section{Conclusion} \label{sec6}
% chart pattern의 의의를 다시 언급
% chart pattern들의 매치 정보는 trading decision을 결정하기 위한 중요한 정보를 제공하는 feature로 풀이될 수 있다. 또는 chart patterns의 조합 그 자체로 trading strategy로 연결될 수도 있다. chart pattern의 분석은 실용적인 가치 반하여 
Attractive chart patterns can provide crucial information for making trading decisions, but the chart pattern search has been relatively unexplored so far.
In this paper, we defined and formulated the problem of chart pattern search which is not limited by matching techniques, and proposed HyperNEAT framework to cope with the problem. We used various deep neural network techniques to overcome problem-specific difficulties. 
Although the neural network-based patterns were better than the others, they were fairly simple. We think that our model is still not mature enough, hoping to find more complex patterns hidden in the pattern space.\\
% The proposed framework successfully found profitable neural network-based chart patterns. We ascertained that the patterns found are rather simple and explainable by observing the matched charts by the patterns.
% Overall, the patterns were superior to the other patterns designed in different ways.\\
In the future, we expect to increase profitability by injecting contextual information such as time or the stock item into the chart.
We also plan to devise a method to speed up the computation of CPPN using GPGPU, and design a new CPPN so that NEAT can adaptively determine the weight scale using information about incoming connections.

\section*{Acknowledgement}
The authors would like to thank Seung-kyu Lee for constructive discussion and proof reading the article. The Institute of Engineering Research of Seoul National University provided research facilities for this work.

%\bibliographystyle{ACM-Reference-Format}
%\bibliography{tbcp}

\end{document}